\documentclass{article}

\usepackage{arxiv}
\usepackage[utf8]{inputenc} 
\usepackage[T1]{fontenc}    
\usepackage{hyperref}       
\usepackage{url}            
\usepackage{booktabs}       
\usepackage{amsfonts}       
\usepackage{nicefrac}       
\usepackage{microtype}      
\usepackage{lipsum}		
\usepackage{graphicx}
\usepackage{doi}
\usepackage{soul, color}
\usepackage{amsmath}
\usepackage{multirow}
\usepackage{adjustbox}

\usepackage{subfig}

\usepackage{algpseudocode}
\usepackage{algorithm}
\algnewcommand\algorithmicforeach{\textbf{for each}}
\algdef{S}[FOR]{ForEach}[1]{\algorithmicforeach\ #1\ \algorithmicdo}

\usepackage[
backend=biber,
style=numeric,
citestyle=nature,
]{biblatex}

\title{Accurate online training of dynamical spiking neural networks through Forward Propagation Through Time}

\author{
  Bojian Yin\\
  CWI \\
  \texttt{Bojian.Yin@cwi.nl} \\
   \And
Federico Corradi \\
  IMEC\\
  \texttt{Federico.Corradi@imec.nl} \\
  \And
  Sander M. Boht\'e\\
  CWI \\
  \texttt{S.M.Bohte@cwi.nl} \\
}

\renewcommand{\shorttitle}{\textit{arXiv} Template}

\hypersetup{
pdftitle={FPTT},
pdfsubject={q-bio.NC, q-bio.QM},
pdfauthor={Bojian Yin, Sander M. Boht\'e},
pdfkeywords={Spiking neural network, FPTT, online learning},
}

\begin{document}

\maketitle

\begin{abstract}
The event-driven and sparse nature of communication between spiking neurons in the brain holds great promise for flexible and energy-efficient AI. Recent advances in learning algorithms have demonstrated that recurrent networks of spiking neurons can be effectively trained to achieve competitive performance compared to standard recurrent neural networks. 
Still, as these learning algorithms use error-backpropagation through time (BPTT), they suffer from high memory requirements, are slow to train, and are incompatible with online learning. This limits the application of these learning algorithms to relatively small networks and to limited temporal sequence lengths. 
Online approximations to BPTT with lower computational and memory complexity have been proposed (e-prop, OSTL), but in practice also suffer from memory limitations and, as approximations, do not outperform standard BPTT training. Here, we show how a recently developed alternative to BPTT, Forward Propagation Through Time (FPTT) can be applied in spiking neural networks. Different from BPTT, FPTT attempts to minimize an ongoing dynamically regularized risk on the loss. As a result, FPTT can be computed in an online fashion and has fixed complexity with respect to the sequence length. When combined with a novel dynamic spiking neuron model, the Liquid-Time-Constant neuron, we show that SNNs trained with FPTT outperform online BPTT approximations, and approach or exceed offline BPTT accuracy on temporal classification tasks. This approach thus makes it feasible to train SNNs in a memory-friendly online fashion on long sequences and scale up SNNs to novel and complex neural architectures. 


\end{abstract}


\section{Introduction}
Recent work has demonstrated effective and efficient performance from spiking neural networks \cite{byin2021AccEffSNN}, enabling competitive and energy-efficient applications in neuromorphic hardware \cite{stuijt2021mubrain} and novel means of investigating biological neural architectures \cite{Perez-Nieves2021-ge,Keijser2020-fg}.
This success stems principally from the use of approximating surrogate gradients \cite{bohte2011error,neftci2019surrogate} to integrate networks of spiking neurons into auto differentiating frameworks like Tensorflow and Pytorch \cite{paszke2019pytorch}, enabling the application of standard learning algorithms and in particular back-propagation through time (BPTT). 

However, the imprecision of the surrogate gradient approach expounds on the existing drawbacks of BPTT. In particular, BPTT has a  linearly increasing memory cost as a function of sequence length $T$, $\Omega(T)$ and can suffer from vanishing or exploding backpropagating gradients, limiting its applicability on long time sequences \cite{kag2021training}. 
Alternative approaches like real-time recurrent learning (RTRL)\cite{williams1989learning} similarly exhibit excessive data-complexity, and 
low time-complexity approximations to BPTT like e-prop \cite{bellec2020solution} or OSTL \cite{bohnstingl2020online} at best approach BPTT performance. In addition, training on long temporal sequences in SNNs is of particular importance when the tasks require a high temporal resolution, for instance to match the physical characteristics of low-latency neuromorphic hardware \cite{stuijt2021mubrain}: as there is no notion of discrete-time steps in clock-less event-driven neuromorphic devices and time is continuous, off-chip SNNs need to be trained on temporal sequences with extremely short time steps to mimic continuous-time characteristics and guarantee corresponding performance in real-life applications~\cite{he2021}.



A novel learning algorithm, Forward~Propagation~Through~Time (FPTT), was recently introduced based on minimizing an instantaneous risk function using dynamic regularization. FPTT was demonstrated to improve long sequence training in Long Short-Term Memory networks (LSTMs) compared to BPTT while exhibiting linear $\Omega(T)$ computational cost per sample. The latter also enables FPTT to learn in an online fashion. As we show, a straightforward application of FPTT to SNNs fails, and we found it similarly failed with standard RNNs: we deduce that FPTT particularly benefits from the gating-structure inherent in LSTMs and GRUs which is lacking in standard RNNs and SRNNs.

Here, and inspired by the concept of Liquid Time Constants (LTCs) \cite{Hasani2020}, we introduce a novel class of spiking neurons, the Liquid Spiking Neuron, where internal time-constants are dynamic and input-driven in a learned fashion, resulting in functionality similar to the gating operation in LSTMs. We then integrate these Liquid Spiking Neurons in SNNs that are trained with FPTT.
 
We demonstrate that LTC-SNNs networks trained with FPTT outperform various SNNs trained with BPTT on long sequences while enabling online learning and drastically reducing memory complexity. We show this for a number of classical benchmarks that can easily be varied in duration, like the adding task and the DVS gesture benchmark ~\cite{amir2017low,fang2021incorporating}.
We also show how LTC-SNNs trained with FPTT can be applied to large-scale convolutional SNNs, where we demonstrate novel state-of-the-art for online learning in recurrent SNNs on several standard benchmarks (S-MNIST, R-MNIST, DVS-GESTURE) and also show that large feedforward SNNs can be trained successfully in an online manner to near state-of-the-art performance as obtained with offline BPTT (Fashion-MNIST, DVS-CIFAR10).



\section{Related Work}

The problem of training recurrent neural networks has an extensive history, including early work by Werbos \cite{werbos1990backpropagation}, Elman \cite{elman1990finding} and Mozer \cite{mozer1993neural}. In a recurrent network, to account for past influences on current activations, the network is unrolled and errors are computed along the paths of the unrolled network. The direct application of error-backpropagation to this unrolled graph is known as Backpropagation-Through-Time \cite{werbos1990backpropagation}. BPTT needs to wait until the last input of a sequence before being able to calculate parameter updates and, as such, cannot be applied in an online manner. Alternative online learning algorithms for RNNs have been developed, including Real-Time Recurrent Learning (RTRL) \cite{williams1989learning} and mixes of both RTRL and BPTT approaches \cite{murray2019local}; they however exhibit prohibitive time and memory complexity \cite{bohnstingl2020online}. See Table \ref{tab:FPTT_complex} for overview.

For networks of spiking neurons, the discontinuity of the spiking mechanism challenges the application of error-backpropagation, which can be overcome using continuous approximations~\cite{bohte2002error,bohte2011error}, so-called ``surrogate gradients
'' \cite{neftci2019surrogate}. Various SNNs trained with such surrogate gradients and BPTT now achieve competitive performance compared to classical RNNs \cite{yin2020effective,byin2021AccEffSNN,fang2021incorporating}. 
While effective, the application of BPTT in SNNs has several drawbacks: 
in particular, BPTT accumulates the approximation error of surrogate gradients along time. Furthermore, because the SNN performance heavily depends on hyperparameters related to the surrogate gradients, obtaining convergence in SNN networks is non-trivial.
Moreover, the spike-triggered reset of the membrane potential due to refraction causes a vanishing gradient. 


Approximations to BPTT like e-prop \cite{bellec2020solution} achieve linear time complexity and have proven effective for many small scale benchmark problems and also large scale networks like cortical microcircuits \cite{ScherrBioRXiv2021}. Online Spatio-Temporal Learning (OSTL) \cite{bohnstingl2020online} separates the spatial and temporal gradient calculations to derive weight updates in an online manner, but suffers from very high computational and memory costs for generic RNNs. Still, in terms of trained accuracy, none of these approximations have been shown to outperform standard BPTT. 


\begin{table}[t]
\centering
\caption{Computational complexity of gradients, parameter updates and memory storage per  sample. The computational expense increases as the length of the sequence grows. i.e. $c(1)<c(T)$. After \cite{kag2021training}.}
\small
\begin{tabular}{|l|l|l|l|}
\hline
Algorithm & \begin{tabular}[c]{@{}l@{}}Gradient\\ Update\end{tabular} & \begin{tabular}[c]{@{}l@{}}parameter\\ Update\end{tabular} & \begin{tabular}[c]{@{}l@{}}Memory\\ Storage\end{tabular} \\ \hline
BPTT      &             $\Omega(c(T)T)$                               &   $\Omega(1)$                                              & $\Omega(T)$                                               \\ \hline
RTRL      &             $\Omega(c(T)T^2)$                             &   $\Omega(T)$                                              & $\Omega(T)$                                              \\ \hline
e-prop / OSTL    &             $\Omega(c(1)T)$                               &  $\Omega(T)$                                               & $\Omega(1)$                                                \\ \hline
FPTT      &             $\Omega(c(1)T)$                               &  $\Omega(T)$                                               & $\Omega(1)$                                              \\ \hline
FPTT-K    &             $\Omega(c(K)T)$                               &  $\Omega(K)$                                               & $\Omega(T/K)$                                              \\ \hline
\end{tabular}
\label{tab:FPTT_complex}
\end{table}

\begin{figure}[t]
	\centering
	\includegraphics[width=0.8\textwidth]{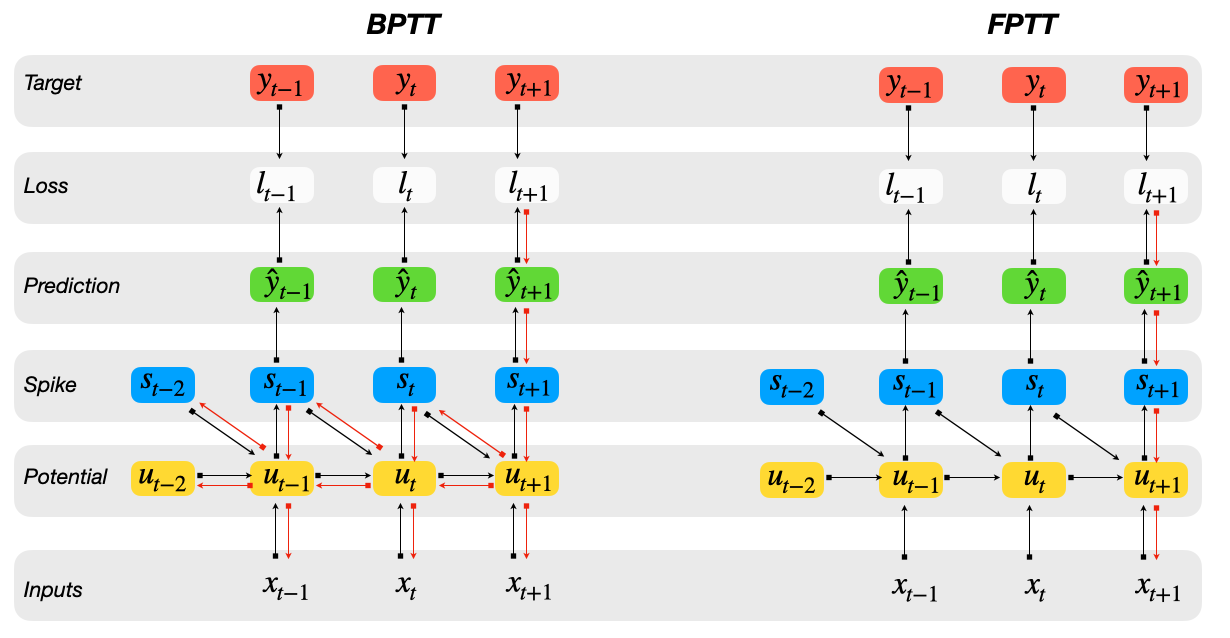}
	\caption{Roll-out of the computational graph of a spiking neuron as used for BPTT (left) and that of FPTT (right).}
	\label{fig:bptt_fptt}
\end{figure}

\paragraph{FPTT.} Forward Propagation through Time (FPTT) \cite{kag2021training} updates the network parameters by optimising the instantaneous risk function $\ell_t^{dyn}$, which includes the ordinary objective $\mathcal{L}_t$ and also a dynamic regularisation penalty $\mathcal{R}_t$ based on previously observed losses $\ell_t^{dyn} = \mathcal{L}_t+ \mathcal{R}_t$ (see Appendix A for details). By adding this dynamically time-evolving regularizer, FPTT optimizes RNNs similar to feedforward networks, as shown in the computational diagram of Fig.\ref{fig:bptt_fptt}. FPTT thus eliminates the dependence of the gradient calculation on the sum of products of partial gradients along the time dimension in BPTT.

In detail, first, the empirical objective $\mathcal{L}(y_t,\hat{y}_t)$ is the same as that of BPTT, representing the gap between target values $y_t$ and real time predictions $\hat{y}_t)$. Second, and most important, the novel dynamic regularization part is controlled by the ``running average'' of all the weights seen so far. The update schema of this regularizer is as follows:
\begin{equation}
\begin{split}
\mathcal{R}(\bar{\Phi}_t) & = \frac{\alpha}{2}\parallel \Phi - \bar{\Phi}_{t} - \frac{1}{2\alpha}\nabla l_{t-1}(\Phi_t) \parallel\\
\Phi_{t+1} &= \Phi_t - \eta\nabla_{\Phi}l(\Phi)|_{\Phi=\Phi_t}\\
\bar{\Phi}_{t+1}&= \frac{1}{2}(\bar{\Phi}_t+\Phi_{t+1})-\frac{1}{2\alpha}\nabla l_t(\Phi_{t+1})
\end{split}
\end{equation}
In FPTT, a state vector $\bar{\Phi}_t$ is introduced which summarises past losses: the first update is a normal update of parameters $\Phi_t$ based on gradient optimization with fixed $\bar{\Phi}_t$; after the update, we optimize $\bar{\Phi}_t$ with fixed $\Phi_t$. This approach allows RNN parameters to converge to a stationary solution of the traditional RNN objective \cite{kag2021training}. 

The FPTT learning process requires the acquisition of an instantaneous loss $l_t$ at each time step. This is natural for sequence-to-sequence modelling tasks and streaming tasks where a loss is available for each time step; for classification tasks, however, the target value is only determined after processing the entire time series. To adapt FPTT to classification tasks, or rather, to perform online classification tasks, Kag \& Saligrama \cite{kag2021training} introduced a divergence term in the form of an auxiliary loss to reduce the distance between the prediction distribution $\hat{P}$ and target label distribution $Q$:
\begin{equation}
    l_t = \beta l_t^{CE}(\hat{y}_y,y) + (1-\beta)l_t^{div},
\end{equation}
where $\beta \in [0,1]$; $l_t^{CE}$ is the classical cross-entropy for a classification loss and $l_t^{div} = -\sum_{\bar{y}}Q(\bar{y})\log \hat{P}(\bar{y})$ is the divergence term.

\section{Training networks of spiking neurons} 
To apply FPTT to SNNs, we first define the spiking neuron model and explain how BPTT is applied to such networks. An SNN is comprised of spiking neurons which operate with non-linear internal dynamics. These non-linear dynamics consist of three main components: 

\textbf{(1) Potential Updating:} the neurons' membrane potential $u_{t}$ updates following the equation: 
\begin{equation}
\label{eq:mem_update}
u_{t} = f(u_{t-1},x_t,s_{t-1}\| \Phi,\tau)
\end{equation}
where $\tau$ is the set of internal time constants and $\Phi$ is the set of associated parameters like synaptic weights. The membrane potential evolves along time based on previous neuronal states (e.g. potential $u_{t-1}$ and spike-state $s_{t-1}$) and current inputs $x_t$. 

\textbf{(2) Spike generation:} A neuron will trigger a spike $s_t=1$ when its membrane potential $u_t$ crosses a threshold $\theta$ from below, described as a discontinuous function:
\begin{equation}
\label{eq:f_spike}
s_t = f_s(u_t,\theta) = 
\begin{cases}
    1,& \text{if } u_t\geq \theta\\
    0,              & \text{otherwise}
\end{cases}
\end{equation}

\textbf{(3) Potential resting:} When it emits a spike ($s_t = 1$), the membrane potential will reset to resting potential $u_{r}$. In all experiments, we set $u_{r}=0$:
\begin{equation}
\label{eq:mem_reset}
u_t = (1-s_t)u_t+u_{r} s_t
\end{equation}
For optimal performance, the various time constants in the spiking neurons can be learned to match the temporal dynamics of the task \cite{yin2020effective,fang2021incorporating}. 


\paragraph{BPTT for SNNs.} As outlined in Algorithm 1, BPTT for SNNs amounts to the following: given a training example $\{x,y\}$ of T time steps, the SNN generates a prediction $\hat{y_t}$ at each time step.  At time $t$, the SNN parameters are optimized by gradient descent through BPTT to minimize the instantaneous objective $\ell_t = \mathcal{L}(y_t,\hat{y}_t)$. The gradient expression is the sum of the products of the partial gradients, defined by the chain rule as 
\begin{equation}
\label{eq:bptt_eq}
\frac{\partial \ell_{t+1}}{\partial \Phi} = \frac{\partial l_{t+1}}{\partial \hat{y}_{t+1}} \frac{\partial \hat{y}_{t+1}}{\partial s_{t+1}} \frac{\partial s_{t+1}}{\partial u_{t+1}} \sum_{j=1}^{t+1}\left(\prod_{m=j}^{t+1} \frac{\partial u_{m}}{\partial u_{m-1}}\right) \frac{\partial s_{m-1}}{\partial \Phi}
\end{equation}

where the partial derivative of spike $\frac{\partial s_t}{\partial u_t}$ is calculated by a surrogate gradient associate with membrane potential $u_t$. Here, we use the Multi-Gaussian surrogate gradient function $\hat{f'}_s(u_t,\theta)$ \cite{byin2021AccEffSNN} to approximate this partial term. 

The computational graph of BPTT is shown in Fig\ref{fig:bptt_fptt} and shows that the partial derivative term depends on two pathways, $\frac{\partial u_{m}}{\partial u_{m-1}} = \frac{\partial u_{m}}{\partial u_{m-1}}+\frac{\partial u_{m}}{\partial s_{m-1}}\frac{\partial s_{m-1}}{\partial u_{m-1}}$. This gradient expression implies that the error of the surrogate gradient accumulates and amplifies during the training process. The product of these partial terms may explode or vanish in RNNs, and this phenomenon becomes even more pronounced in SNNs.

\paragraph{FPTT for SNN} FPTT can be used for training SNNs as described in Algorithm 2: we optimize the network by minimizing the instantaneous loss with the dynamic regularizer $\ell_t^{dyn} = \mathcal{L}(y_t,\hat{y}_t)+ \mathcal{R}(\bar{\Phi}_t)$. For FPTT, the update function Equation \eqref{eq:bptt_eq} then becomes: 
\begin{equation}
\frac{\partial \ell_{t+1}^{dyn}}{\partial \Phi} = \frac{\partial l_{t+1}}{\partial \hat{y}_{t+1}} \frac{\partial \hat{y}_{t+1}}{\partial s_{t+1}} \frac{\partial s_{t+1}}{\partial u_{t+1}}  \frac{\partial u_{t+1}}{\partial \Phi}
\label{eq:fptt_eq}
\end{equation}
Compared to Equation \eqref{eq:bptt_eq}, Equation \eqref{eq:fptt_eq} has no dependence on a chain of past states, and can thus be computed in an online manner. Theoretically, FPTT provides a more robust and efficient gradient approximation for recurrent neural networks to avoid gradient vanishing or explosion. For SNNs, FPTT simplifies the complex gradient computation path in BPTT, potentially weakening the effect of surrogate gradients and providing a better gradient approximation for the network. 

\begin{algorithm} 
\caption{Training SNN with BPTT}
\begin{algorithmic}[1]
\Require $B = \{x_t,y_t\}_{t=0}^T$, \# Epochs $E$
\Require Optimizer and learning rate $\eta$
\State Initialize Weight $W,v$, 
\ForEach {$e \in E $}
    \State Initialize Neuron states $u_t, s_t$
    \State Randomly Shuffle $B$
    
    \ForEach {$t \in T $}
        \State Update: $s_{h,t},u_{h,t} = \hat{f}_s(x_{t-1},[u_{h,t-1},s_{h,t-1}]\|W)$
        \State Predict: $\hat{y}_t=\hat{f}_s(s_{h,t},[u_{o,t},s_{o,t}]\|v)$
    \EndFor
    
    \State $Loss$: $\ell(W) = \sum_{t=1}^T \ell(y_t,\hat{y}_t)$
    \State $Update$: $W = W-\eta \nabla_W \ell(W)|_W$
\EndFor
\end{algorithmic}
\end{algorithm}
\begin{algorithm} 
\caption{Training SNN with FPTT}
\begin{algorithmic}[1]
\Require $B = \{x_t,y_t\}_{t=0}^T$, \# Epochs $E$
\Require Optimizer and learning rate $\eta$
\State Initialize Weight $W$, and $\bar{W}=W$ 

\ForEach {$e \in E $}
\State Initialize Neuron states $u_t, s_t$
\State Randomly Shuffle $B$

\ForEach {$t \in T $}

\State Update: $s_{h,t},u_{h,t} = \hat{f}_s(x_{t-1},[u_{h,t-1},s_{h,t-1}]\|W)$
\State Predict: $\hat{y}_t=\hat{f}_s(s_{h,t},[u_{o,t},s_{o,t}]\|v)$

\State Loss $\ell_t(W) $: $\ell_t(W) = \ell(y_t,\hat{y}_t)$
\State Dynamic Loss: $\ell^{dyn}(W) = \ell_t(W) + \frac{\alpha}{2}\| W-\bar{W}_t-\frac{1}{2\alpha} \nabla\ell_{t-1}(W_t) \|^2$
\State Update $W$: $W_{t+1} = W_t-\eta \nabla_W \ell(W)|_{W=W_t}$
\State Update $\bar{W}$: $\bar{W}_{t+1} = \frac{1}{2}(\bar{W}_t+W_{t+1})-\frac{1}{2\alpha} \nabla \ell_t(W_{t+1})$
\EndFor
\EndFor
\end{algorithmic}
\end{algorithm}

%

\section{The Liquid Spiking Neuron}

\begin{figure}[ht]
	\centering
	\includegraphics[width=0.3\textwidth]{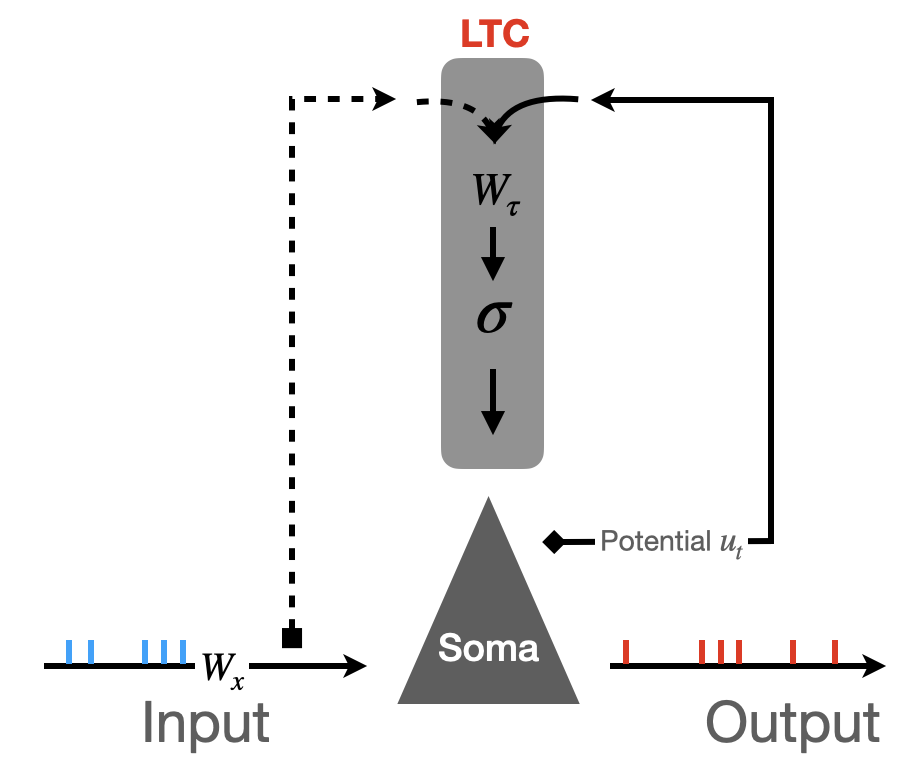}
	\caption{Circuit of Liquid time-constant spiking neuron}
	\label{fig:ltc}
\end{figure}
We here introduce the Liquid Spiking Neuron (LSN) model, which, as we will show, enables the application of FPTT to SNNs. We observe that to some degree the time-constant of the membrane potential acts similar to the forget-gate in LSTMs; the LSTM forget-gate, however, is dynamically controlled by learned inputs. Inspired also by the work by \cite{Hasani2020}, we introduce a spiking neuron model where some of the time-constants are learned functions of the inputs and hidden states of the network, as illustrated in Fig. \ref{fig:ltc}.

Mathematically, we describe a Liquid Time Constant as: 
\begin{equation}
\tau^{-1} = \sigma(x_t,u_{t-1}|W_{\tau}),
\label{eq: ltc}
\end{equation}
where, to ensure smooth changes when learning, we use a sigmoid function to scale the inverse of the time-constant to a range of 0 to 1. 
In detail, the liquid time-constants in standard adaptive spiking neurons \cite{bellec2020solution,byin2021AccEffSNN} are either calculated as a function $\tau_m^{-1} = \sigma(Dense([x_t,u_{t-1}]))$, for non-convolutional networks, or using a 2D convolution for spiking convolutional networks, $\tau_m^{-1}=\sigma(Conv(x_t+u_{t-1}))$. This results in a Liquid Spiking Neuron defined by the following equations:
\begin{equation}
\begin{split}
\begin{aligned}
\tau_{adp} \, update :&  \rho = \tau_{adp}^{-1}=\sigma([x_t,b_{t-1}]\parallel W_{\tau_{adp}}) \\
\tau_m \, update :&  \tau_m^{-1}=\sigma([x_t,u_{t-1}]\parallel W_{\tau_m}) \\
\theta_t \, update :& b_t = \rho b_{t-1}+(1-\rho)s_{t-1}\\
 & \theta_t = 0.1+1.8 b_t \\
u_t \,update :&  du = (-u_{t-1}+x_t)/\tau_m \\
 & u_t = u_{t-1}+du \\
spike \, s_t :& s_t = f_s(u_t,\theta)\\
resting :& u_t = u_t(1-s_t)+u_{rest}s_t,
\end{aligned}
\end{split}
\end{equation}
where the neuron uses an adaptive threshold $\theta_t$ as in the Adaptive Spiking Neurons \cite{bellec2020solution}, and $\tau_m$ and $\tau_m$ are computed as liquid time-constants.

\section{Experiments}

\paragraph{Datasets.} We demonstrate the effectiveness of the FPTT algorithm with LTC-SNNs on a number of classical benchmarks either to compare to \cite{kag2021training} (the Add-task), or to compare to established SNN benchmarks (the DVS Gesture and DVS-CIFAR10 classification tasks, and the Sequential, Sequential-Permuted, rate-based and Fashion MNIST classification tasks).

The {\bf Add-Task} \cite{hochreiter1997long} is used to evaluate the ability of RNNs to maintain long-term memory. An example data point consists of two sequences $(x_1, x_2)$ of length $T$ and a target label $y$. The sequence $x_1$ contains real-valued items sampled uniformly from [0, 1], $x_2$ is a binary sequence of only two $1$s, and the label $y$ is the sum of the two entries in the sequence $x_1$, where $x_2=1$. 
The trained networks consist of 128 recurrently connected neurons of respective types LTC-SNN (LTC-SRNN), LSTM, or Adaptive Spiking Neuron \cite{bellec2020solution} (ASRNN), and a dense output layer with only 1 neuron.

The {\bf IBM DVS Gesture} dataset \cite{amir2017low} consists of 11 kinds of hand and arm movements of 29 individuals under three different lighting condition captured using a DVS128 camera. Each frame is a 128-by-128 size image with 2 channels. {\bf{DVS-CIFAR10}} is a widely used neuromorphic vision dataset where the event stream obtained by displaying the moving images of the CIFAR-10 dataset \cite{li2017cifar10}. As in \cite{fang2021incorporating}, for both DVS-Gesture and DVS-CIFAR10, we cluster the event flow into frames. Since sequence length depends on sampling frequency, we sampled such as to yield various sequence lengths from 20 to 500 frames. 



In the DVS-Gesture dataset, we apply either a shallow spiking recurrent network (SRNN) or a deep spiking convolutional network (SCNN) to test the training effectiveness of FPTT on longer sequences as well as larger and deeper networks. As input for the shallow SRNN, we first down-sample the frame of a 128-by-128 image into a 32-by-32 image by averaging each 4-by-4 pixel block. Then, the 2D image at each channel is flattened into a 1D vector of length 1024. For each channel of the image, the network consists of a spike-dense layer consisting of 512 neurons as an encoder, where the information of each channel is then fused into a 1D binary vector through concatenation. This fused information is then fed to a recurrently connected layer with 512 hidden neurons. Finally, a leaky integrator is applied to generate predictions:  [1024,1024]-[512D,512D]-512R-11I.

To achieve high performance on the DVS Gestureand DVS-CIFAR10 datasets, we follow \cite{fang2021incorporating} and use 20 sequential frames, where the network makes a prediction only after reading the entire sequence. We use a spiking convolutional network following the structure:  ConvK7C64S1P3-MPK2S2-ConvK7C128S1P3-MPK2S2-ConvK3C128S1P1-MPK2S2-ConvK3C256S1P1-MPK2S2-ConvK3C256S1P1-MPK2S2-ConvK3C512S1P1-MPK2S2-512D-11I. The network was optimized through Adamax \cite{kingma2014adam} with a batch size of 16 and initial learning rate of 1e-3. 

The {\bf Sequential and Permuted-Sequential} MNIST (S-MNIST, PS-MNIST) datasets were developed to measure sequence recognition and memory capabilities of learning algorithms. A grey input image of shape 28-by-28 is reshaped into a one-dimensional sequence consisting of 784 time steps. At each time step, only one pixel entered the network as an input. The permuted-MNIST dataset is generated by performing a fixed permutation on the sequential MNIST dataset. Theoretically and in practice, PS-MNIST is more difficult than S-MNIST because it lacks temporally correlated patterns. 
In (P)S-MNIST, we applied a shallow network with one recurrent layer comprised of 512 hidden neurons, and the output layer consists of 10 (number of classes) leaky integrator neurons. Networks are optimized using Adam \cite{kingma2014adam} with a batch size of 128 using 200 training epochs. We set the initial learning rate to 3e-3 and decay by half after 30, 80 and 120 epochs.

The {\bf rate-coded MNIST} (R-MNIST) is an SNN specific benchmark where a biologically inspired encoding method is used to generate the network input that produces streaming events (a spike train) by encoding the grey values of the image with Poisson rate-coding \cite{gerstner1997neural}. As in  \cite{bohnstingl2020online}, we apply an SNN with two hidden layers of 256 neurons each followed by 10 output neurons. The SNN is given 20 presentations of the image, after which the classification is determined. 


We also tested FPTT-trained LTC-SCNNs on the traditional static {\bf MNIST} and {\bf Fashion-MNIST} datasets for comparison with other models trained offline. Pixel values are directly injected as current into the first spiking layer of the network, repeated 20 times to mimic a constant input stream. We apply an SCNN with 3 convolutional layers, 1 Dense layer and 1 leaky Integrator output layer: ConvK3C32S1P1-MPK2S2-ConvK3C128S1P1-MPK2S2-ConvK3C256S1P1-MPK2S2-512D-10I. The network was optimized by Adamax with a batch size of 64 and an initial learning rate of 1e-3.

\begin{figure}[t]%
    \centering
    \subfloat[\centering ]{{\includegraphics[width=0.46\textwidth]{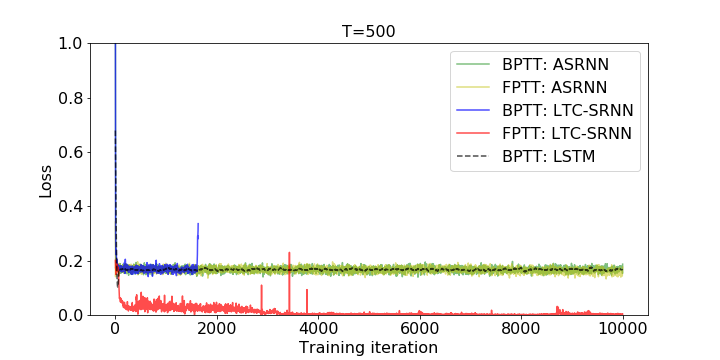} }}%
    \qquad
    \subfloat[\centering ]{{\includegraphics[width=0.46\textwidth]{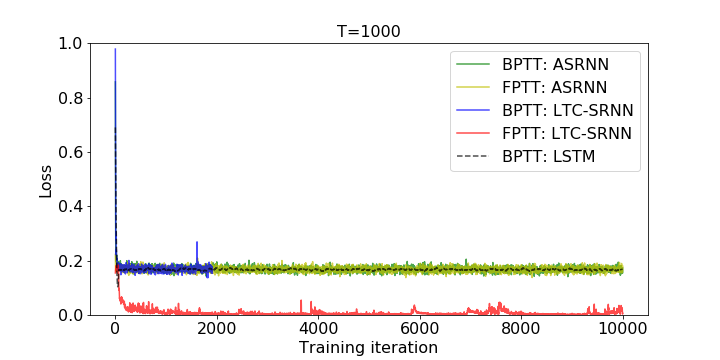} }}%
    
    \subfloat[\centering ]{{\includegraphics[width=0.46\textwidth]{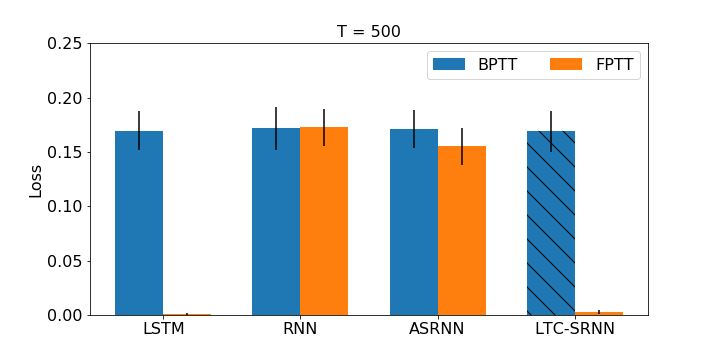} }}%
    \qquad
    \subfloat[\centering ]{{\includegraphics[width=0.46\textwidth]{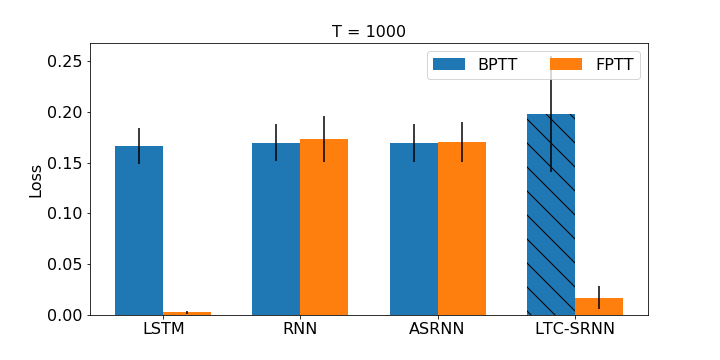} }}%
    
    \caption{Add Task. \textbf{The top row} plots the loss curve for the Add Task for different sequence lengths (a) $T=500$ and (b) $T=1000$. We take as baseline performance a non-spiking LSTM with the same number of neural units trained with BPTT.  \textbf{The second row} plots the loss of the last 100 training iterations averaged over 5 runs for sequence lengths (c) $T=500$ and (d) $T=1000$. The loss of LTC-SRNN becomes $NaN$ after 2000 iterations when training with BPTT. We then report the loss right before divergence, indicated by a hatched texture in the respective bars.}%
    \label{fig:adding_task}%
\end{figure}

\subsection{FPTT-SNN requires Liquid Spiking Neurons}

To illustrate the effectiveness of the FPTT-LTC-SNN framework and the need for liquid time-constant spiking neurons when applying FPTT learning, we apply both BPTT and FPTT to the Add Task. This has been done using various networks, including non-spiking LSTMs as a baseline as in \cite{kag2021training}, adaptive SRNNs (ASRNNs) as in \cite{byin2021AccEffSNN}, and LTC-SRNNs. 

Example loss-curves are shown in Fig. \ref{fig:adding_task}(a,b), for adding sequences of length 500 and 1000. As reported in \cite{kag2021training}, standard LSTMs trained with BPTT do not converge, while they do when applying FPTT training. For SNNs, we find that ASRNNs as in \cite{byin2021AccEffSNN} trained with either FPTT or BPTT do not converge; for LTC-SNNs trained with BPTT, we find that the loss initially decreases to values similar to that of standard SRNNs, but then learning diverges due to exploding gradients. Finally, LTC-SNNs trained with FPTT successfully minimize the loss similar to the FPTT-LSTM networks. The final losses averaged over 5 networks are shown in Fig \ref{fig:adding_task}(c,d).

\subsection{FPTT allows for longer sequence training}
We next study the ability of FPTT-style training of LTC-SNNs to learn increasingly long sequences. With the DVS Gesture dataset, we systematically investigate the performance of shallow SRNN models on increasingly many frames of high frequency sampled signals: we converte the entire event stream into sequences of differing lengths, ranging from 20 to 500 frames. The longer sequence lengths pose a serious challenge to a network's ability to memorize relevant information at different time scales. Furthermore, it also places increasingly high demands on memory and training time for training via BPTT.

We examine the performance of various network architectures, training methods and loss functions. In particular, we trained a set of networks with an identical number of neural units: LSTM networks with BPTT, with and without the auxiliary loss, and similarly trained ASRNNs and recurrent LTC-SNNs (LTC-SRNNs). This we compared to LTC-SRNNs trained with FPTT with the auxiliary loss. The results for different sampling frequencies are listed in Table \ref{tab:dvs_fptt_bptt}.

\begin{table}[h!t]
\centering
\caption{\textbf{Performance comparison between BPTT and FPTT on the DVS gesture dataset.} Each number in the table is the average of three runs. All networks have equal number of neural units.(*): training diverged; reported accuracy is best accuracy before divergence.\vspace{0.2cm}}
\small
\begin{adjustbox}{max width=\textwidth}
\begin{tabular}{|c|cccccc|}
\hline
\multirow{2}{*}{Frames}                              
                        & \multicolumn{5}{l|}{BPTT}                                                                                              & FPTT  \\ \cline{2-6}
                        
                        & \multicolumn{1}{l|}{LSTM+Aux} & \multicolumn{1}{l|}{LSTM}  & \multicolumn{1}{l|}{LTC-SRNN+Aux} & \multicolumn{1}{l|}{LTC-SRNN} &  \multicolumn{1}{l|}{ASRNN+Aux}  & LTC-SNN   \\ \hline
20                      & \multicolumn{1}{l|}{86.69$\pm$0.43}    & \multicolumn{1}{l|}{82.29$\pm$2.46} & \multicolumn{1}{l|}{83.42$\pm$1.35}   & \multicolumn{1}{l|}{84.37$\pm$2.27} &\multicolumn{1}{l|}{79.16$\pm$1.98}& \textbf{88.31}$\pm$0.59    \\ \hline
40                      & \multicolumn{1}{l|}{88.77$\pm$1.71}    & \multicolumn{1}{l|}{84.95$\pm$0.71} & \multicolumn{1}{l|}{85.96$\pm$1.16}   & \multicolumn{1}{l|}{84.37$\pm$1.24} &\multicolumn{1}{l|}{80.78$\pm$1.40}& \textbf{90.39}$\pm$0.71     \\ \hline
60                      & \multicolumn{1}{l|}{87.61$\pm$0.86}    & \multicolumn{1}{l|}{85.15$\pm$0.75} & \multicolumn{1}{l|}{85.62$\pm$1.18}   & \multicolumn{1}{l|}{83.91$\pm$0.71} &\multicolumn{1}{l|}{80.55$\pm$0.49}& \textbf{90.74}$\pm$0.16        \\ \hline
80                      & \multicolumn{1}{l|}{87.97$\pm$0.14}    & \multicolumn{1}{l|}{84.83$\pm$1.42} & \multicolumn{1}{l|}{85.30$\pm$0.71}   & \multicolumn{1}{l|}{80.44$\pm$3.6} &\multicolumn{1}{l|}{76.04$\pm$0.85}& \textbf{91.31}$\pm$0.98       \\ \hline
100                     & \multicolumn{1}{l|}{88.89$\pm$0.49}    & \multicolumn{1}{l|}{83.79$\pm$0.71} & \multicolumn{1}{l|}{83.21$\pm$0.43}   & \multicolumn{1}{l|}{78.70$\pm$0.91} &\multicolumn{1}{l|}{74.3$\pm$0.84}& \textbf{91.89}$\pm$0.16       \\ \hline
200                     & \multicolumn{1}{l|}{85.76$\pm$0.49}    & \multicolumn{1}{l|}{81.87$\pm$2.58} & \multicolumn{1}{l|}{51.39$\pm$6.0}   & \multicolumn{1}{l|}{43.98$\pm$2.35} &\multicolumn{1}{l|}{64.87$\pm$0.78}& \textbf{90.16}$\pm$1.43       \\ \hline
500                     & \multicolumn{1}{l|}{82.52$\pm$1.82}    & \multicolumn{1}{l|}{78.81$\pm$1.5} & \multicolumn{1}{l|}{38.89$\pm$3.22}   & \multicolumn{1}{l|}{36.46$\pm$1.5} &
\multicolumn{1}{l|}{48.32$\pm$2.0(*)}& \textbf{90.64}$\pm$1.56   \\ \hline
\end{tabular}
\end{adjustbox}
\label{tab:dvs_fptt_bptt}
\end{table}

\begin{table}[h!t]
\centering
\caption{\textbf{Firing rate (fr) and training-time-per-frame comparison between BPTT and FPTT on the DVS gesture dataset.} Each number in the table is the average of three runs. \vspace{0.2cm}}
\small
\begin{tabular}{|c|cc|c|cc|}
\hline
\multirow{3}{*}{Frames}   & \multicolumn{3}{l|}{Fr}                     & \multicolumn{2}{l|}{LTC-SNN Time (s)}                                        \\ \cline{2-6} 
         & \multicolumn{2}{l|}{BPTT}             & FPTT   & \multicolumn{1}{l|}{\multirow{2}{*}{FPTT}} & \multirow{2}{*}{BPTT} \\ \cline{2-4}
                        
        & \multicolumn{1}{l|}{SRNN+Aux} & LTC-SRNN    & SRNN    & \multicolumn{1}{l|}{}                      &                       \\ \hline
20      & \multicolumn{1}{l|}{0.242}  & 0.255 & 0.206  & \multicolumn{1}{l|}{0.4}                     & 0.75                    \\ \hline
40      & \multicolumn{1}{l|}{0.202}  & 0.220 & 0.163 & \multicolumn{1}{l|}{0.36}                    & 1.13                   \\ \hline
60      & \multicolumn{1}{l|}{0.173}  & 0.198 & 0.138 & \multicolumn{1}{l|}{0.32}                    & 1.67                  \\ \hline
80      & \multicolumn{1}{l|}{0.142}  & 0.171 & 0.126 & \multicolumn{1}{l|}{0.29}                    & 2.00                   \\ \hline
100     & \multicolumn{1}{l|}{0.120}  & 0.144 & 0.119 & \multicolumn{1}{l|}{0.33}                    & 2.50                    \\ \hline
200     & \multicolumn{1}{l|}{0.089}   & 0.088 & 0.091 & \multicolumn{1}{l|}{0.38}                    & 4.56                   \\ \hline
500     & \multicolumn{1}{l|}{0.040}  & 0.076 & 0.071 & \multicolumn{1}{l|}{0.4}                   & 11.4                   \\ \hline
\end{tabular}
\label{tab:dvs_fptt_bptt_fr}
\end{table}

From the Table \ref{tab:dvs_fptt_bptt}, we first observe that the LTC-SRNN trained using FPTT achieved the best performance in all cases, also outperforming standard (BPTT-trained) LSTMs. The FPTT-trained LTC-SRNN moreover exhibits essentially constant performance over the whole range of sequence lengths. In contrast, the accuracy of both LTC-SRNNs and ASRNNs quickly deteriorate as sequence length increases, from 85.5\% for 60 frames to 38.9\% for 500 frames for the best performing LTC-SRNN.  
For the baseline standard LSTM this effect is also there, albeit more moderate (decreasing from 88.9\% at 100 frames to 82.5\% at 500 frames). This suggests that indeed the gradient approximation errors in SNNs add up when training with BPTT.

\begin{figure}[b]%
    \centering
    \subfloat[\centering Time Efficiency]{{\includegraphics[width=0.43\textwidth]{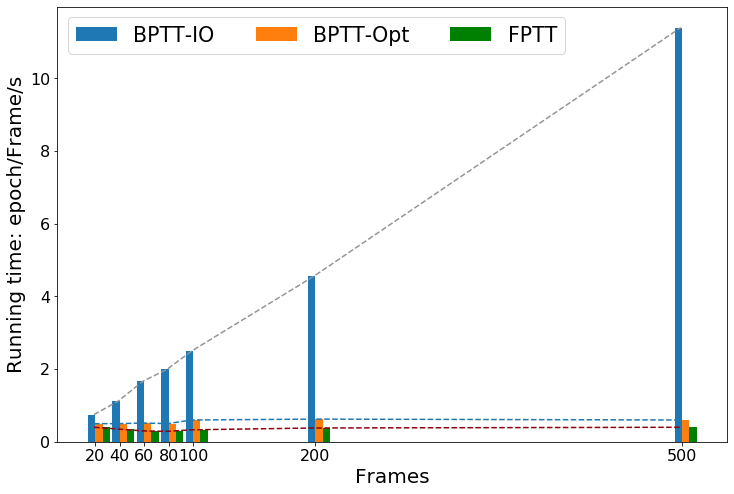} }}%
    \qquad
    \subfloat[\centering Memory Efficiency]{{\includegraphics[width=0.43\textwidth]{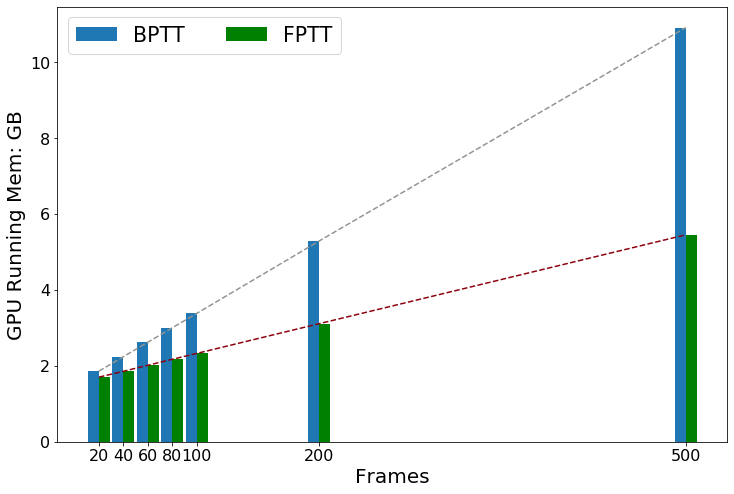} }}%
    \caption{(a) Time Efficiency per frame: training speed of the network using BPTT and FPTT on the DVS-Gesture dataset with different sampling frequencies, where for BPTT both standard (BPTT-Opt) and FPTT-like IO (BPTT-IO) is plotted. (b) Memory Efficiency: Memory cost required for network training of the DVS-Gesture dataset at different sampling frequencies. We set the batch size to 64.}%
    \label{fig:time_mem_eff}%
\end{figure}

We also noted the sparsity (average firing rate) and training time per frame in Table \ref{tab:dvs_fptt_bptt_fr}. For sparsity, we find no meaningful differences between BPTT-SNNs and the FPTT-LTC-SNN; in terms of training time, we find that, when using identical sample-memory-retrieval IO, FPTT-trained LTC-SRNNs train increasingly faster than the BPTT-SNNs; when using the Pytorch optimized BPTT training routine, FPTT was faster consistently by an approximately constant factor (Fig. \ref{fig:time_mem_eff}a).

For on-GPU memory consumption, FPTT-LTC-SNNs require increasingly less memory as sequence length increases (Fig. \ref{fig:time_mem_eff}b). Additionally, for the DVS-Gesture dataset trained on a sequence length of 500 frames, for a batch size of 64 the LSTM-BPTT implementation uses 10.8GB of GPU-memory, the LTC-SNN-BPTT version uses 13.2GB, while the LTC-SNN-FPTT uses 5.6GB. Increasing the batch size to 128 then causes the BPTT versions to no longer fit in our GPU memory (24GB) while the FPTT-LTC-SNN uses 9.6GB.


For both training time and memory usage, we note that the FPTT implementation is unoptimized in the used version of PyTorch, where for instance the memory allocated to historical hidden states is not de-allocated for FPTT, resulting in unnecessary large memory use, and low-level optimized FPTT implementations should further reduce memory to near constant.

\begin{table}[t]
\centering
\caption{Performance of deep SRNNs/SCNNs on various tasks. \vspace{0.2cm}}
\small
\begin{tabular}{c|ccc|cc|c}
\multirow{2}{*}{Task}          & \multicolumn{3}{c|}{\textbf{Online}}                                  & \multicolumn{2}{c|}{\textbf{Offline} {[}BPTT{]}} & \multirow{2}{*}{\begin{tabular}[c]{@{}c@{}}this work\\ LTC-SRNN\end{tabular}} \\ \cline{2-6}
                               & Algorithm          & Network            & test Acc        & Network     & Test Acc               &                                                                              \\ \hline
S-MNIST                        & e-prop\cite{bellec2018long}             & LSNN               &   94.32\%          & ASRNN\cite{byin2021AccEffSNN}       & 98.7\%                   &    97.37\%                                                                          \\ \hline
PS-MNIST                       & -                  & -                  & -                  & ASRNN\cite{byin2021AccEffSNN}        & 94.3\%                   &     94.77\%                                                                        \\ \hline
R-MNIST                        & OSTL\cite{bohnstingl2020online}               &SNU\cite{wozniak2020deep}                 & 95.54\%            &SNU\cite{wozniak2020deep}           & 97.72\%                   &    98.63\%                                                                   \\ \hline

\multirow{2}{*}{MNIST} & \multirow{2}{*}{-} & \multirow{2}{*}{-} & \multirow{2}{*}{-} & LISNN\cite{cheng2020lisnn}       &99.5\%                    & \multirow{2}{*}{99.62\%}                                                            \\ \cline{5-6}
                               &                    &                    &                    & PLIF\cite{fang2021incorporating}         & \multicolumn{1}{c|}{99.72\%}     &                                                                              \\ \hline
\multirow{2}{*}{Fashion-MNIST} & \multirow{2}{*}{-} & \multirow{2}{*}{-} & \multirow{2}{*}{-} & LISNN\cite{cheng2020lisnn}       &92.07\%                    & \multirow{2}{*}{93.58\%}                                                            \\ \cline{5-6}
                               &                    &                    &                    & PLIF\cite{fang2021incorporating}         & \multicolumn{1}{c|}{94.38\%}     &                                                                              \\ \hline

DVS-Gesture                        & DECOLLE\cite{kaiser2020synaptic}               &SNN                 & 95.54\%            &PLIF\cite{fang2021incorporating}          & 97.57\%                   &   97.22\%                                                                   \\ \hline
\multirow{2}{*}{DVS-CIFAR10} & \multirow{2}{*}{-} & \multirow{2}{*}{-} & \multirow{2}{*}{-} & SNN\cite{wu2019direct}       &  60.5\%                         & \multirow{2}{*}{72.3\%}                                                            \\ \cline{5-6}
                               &                    &                    &                  & PLIF\cite{fang2021incorporating}        & \multicolumn{1}{c|}{74.8\%}     &                                                                              \\ \hline
\end{tabular}
\label{tab:result}
\end{table}

\subsection{FPTT with LTC Spiking Neurons improves over Online BPTT}
We further compare large and deep LTC-SCNNs trained with FPTT on standard benchmarks. 
As shown in Table \ref{tab:result}, we find that LTC-SCNNs trained with FPTT consistently outperform SNNs trained with online BPTT approximations like OSTL and e-Prop. Compared to offline BPTT approaches, the online FPTT-trained LTC-SRNNs achieve new SoTa for SNNs (PS-MNIST, R-MNIST) or achieve close to similar performance (S-MNIST, DVS-Gesture, DVS-Cifar10). 




For these large networks, we also find that the memory requirements for FPTT is substantially lower than for BPTT training by a factor of 4 to 5 (Table \ref{tab:largeSNNmem}), and training time is substantially reduced, typically by  a factor of 3 to 4 (Table \ref{tab:traintimelarge}).

\begin{table}[!hb]
\centering
\caption{Memory efficiency. (*): Models using the same batch size cannot be trained on a single GPU, the reported number is obtained using a halved batch size. \vspace{0.2cm}}
\small
\begin{tabular}{|l|l|l|l|l|}
\hline
      & S-MNIST & rate-MNIST & MNIST & DVS-Gesture \\ \hline
BPTT  & 11.1GB & 1.5GB            & 9.67GB            & 15.72GB(*)          \\ \hline
FPTT  & 1.9GB  &1.4GB       & 2.23GB         & 3.75GB       \\ \hline
\end{tabular}

\label{tab:largeSNNmem}
\end{table}

\begin{table}[]
\centering
\caption{Total training time. \vspace{0.2cm}}
\small
\begin{tabular}{|l|l|l|l|l|}
\hline
      & S-MNIST & rate-MNIST & MNIST & DVS gesture \\ \hline
BPTT-IO  & -   & 18min       & 25min       &   -        \\ \hline
BPTT-opt& 40min& 192s        & 362s        & 108s          \\ \hline
FPTT  & 737s   &   204s     &    384s      &    112s    \\ \hline
\end{tabular}
\label{tab:traintimelarge}
\end{table}


\newpage
\section{Discussion}
We showed how a novel training approach, FPTT, can be successfully applied to long sequence learning with recurrent SNNs using novel Liquid Spiking Neurons. Compared to BPTT, FPTT is compatible with online training, has constant memory requirements, trains substantially faster even without optimizations in the software framework and can learn longer sequences with a constant network architecture. In terms of accuracy, FPTT substantially outperforms online approximations to BPTT like OSTL and eProp. Additionally, when training large deep SCNNs with FPTT, excellent performance is achieved approaching or exceeding not-online BPTT-based solutions, including a first demonstration of online learning of tasks like DVS-CIFAR10.

To achieve these results, we introduced Liquid Time-Constant Spiking Neurons (LTC-SN), where the time-constants in the neuron are computed as a learned dynamic function of the current state and input. The LTC-SN is inspired by the functioning of pyramidal neurons in brains, where the apical tuft is coupled to somatic processing \cite{Sacramento_undated-vc,Gidon2020-qq}. Pyramidal neurons are known to have complex non-linear interactions between different morphological parts far exceeding the simple dynamics of LIF-style neurons \cite{beniaguev2021single}, where the apical tuft may calculate a modulating term acting on the computation in the soma \cite{Larkum2004-bt}, which could act similar to the trainable Liquid time-constants used in this work. In a similar vein, learning rules derived from weight-specific traces may relate to synaptic tags \cite{frey1997synaptic,moncada2011identification} and are central to biologically plausible theories of learning working memory \cite{rombouts2015attention}.

In terms of improvements in training time and memory use, the benefits of FPTT versus BPTT were substantial, however they were less than theoretically should be the case. We believe that here, the principal cause are the low-level optimizations in frameworks like Pytorch that are implemented for BPTT but not (yet) for FPTT: when we rearranged BPTT to have a similar main memory access pattern as our FPTT implementation, BPTT training time showed a quadratic increase, as expected.  

The use of FPTT may also hold promise for network quantization: FPTT uses both (a form of) synaptic traces in the form of $\bar{W}$ and the actual weights $W$, where the traces are only used for training. One could imagine networks where the two parameter sets are each calculated with different quantizations, where $W$ could potentially be computed with lower precision compared to $\bar{W}$. Once trained, only the lower precision weights $W$ are then needed for inference. When combined with local error-backpropagation solutions like BrainProp \cite{pozzi2020attention}, FPTT-training of LTC-SNNs can also likely be implemented fully locally and online on neuromorphic hardware.


Together, we believe this work suggests that FPTT may be an excellent training paradigm for SNNs, particularly for LTC-SNNs, which introduce the idea of local online updates necessary in biologically constrained neural information processing systems.

\paragraph{Acknowledgement} BY is supported by the NWO-TTW Programme ``Efficient Deep Learning'' (EDL) P16-25, and SB is supported by the European Union (grant agreement 7202070 ``Human Brain Project'').

\printbibliography

\clearpage
\newpage
\appendix
\section*{Appendix A: FPTT theory}
\setcounter{equation}{0}
\renewcommand\theequation{A.\arabic{equation}}

For conciseness, we briefly summarize the theory underlying FPTT as developed by Kag et al.\cite{kag2021training}.

\paragraph{Back-propagation-through-time} Back-propagation-through-time (BPTT) uses backpropagation to calculate the gradient of the accumulated loss along the spatial-temporal dimension with respect to the parameters of the recurrent networks. Let us define a recurrent network described by differential equation $(\hat{y}^t,h_t) = NN(x_t,h_{t-1})$ where $x_t$ is the input , $\hat{y}^t$ is the prediction and $h_t$ is the hidden states. The gradient of time $t$ is then computed by considering the effect of the state $x_t$ on all future losses $l^t,l^{t+1},....l^T$:

\begin{equation} \label{eq:bptt}
\frac{\partial L}{\partial w} = \sum_{t=1}^T \frac{\partial l^t}{\partial w} = \sum_{t=1}^T\sum_{i=1}^t \frac{\partial l^t}{\partial h_i}\frac{\partial h_i}{\partial w} = \sum_{t=1}^T ( \sum_{i=t}^T \frac{\partial l^i}{\partial h_t} ) \frac{\partial h_t}{\partial w} = \sum_{t=1}^T ( \sum_{i=t}^T \frac{\partial l^i}{\partial h_t} )\frac{\partial h_t}{\partial w} = \sum_{t=1}^T \{ \sum_{i=t}^T ( \prod_{j=i}^{T-1} \frac{\partial l^{j+1}}{\partial l^j})\frac{\partial l^i}{\partial h_t} \}\frac{\partial h_t}{\partial w},
\end{equation} 
and a weight is then updated as: 
$w_{new} \leftarrow w_{old} - \frac{\partial L}{\partial w}$. 
At the end of training, the loss $L$ will be minimized via optimal solution $w^*$, where $ \frac{\partial}{\partial w} L(w^*) \cong 0$. 

For online computation,  we will have $w_{t+1} \leftarrow w_{t} - \sum_{i = 1}^t \frac{\partial}{\partial w}  l^i(\hat{y}^i,y^i,w_t)$ where $l^i(\hat{y}^i,y^i,w_t)$ is the cost of time step $i$ with parameter $w_t$, $\hat{y}^i$ and $y^i$ are the prediction and target label of the time step $i$. 
When the algorithm converges to an optimal solution $w^*$ at time step $\varphi$, we will have an optimal solution where:
\begin{equation}
w^* - w_t = - \nabla_w(l^t) = \frac{\partial}{\partial w} l^{\varphi}(\hat{y}^{\varphi},y^{\varphi},w^*)  -  \frac{\partial}{\partial w} l^t(\hat{y}^t,y^t,w_t)
\end{equation}
and, for one step optimization:
\begin{equation} \label{eq:a1}
w_{t+1} - w_t = \nabla_w l^{t+1} - \nabla_w l^{t}.
\end{equation}
This demonstrates that for any timestep, the change of weight update is proportional to the change of the gradient; this observation (Equation \eqref{eq:a1}) is the foundation of Forward Propagation Through Time.
\paragraph{Forward Propagation Through Time}
FPTT aims to derive an online weight update mechanism with guaranteed convergence to optimal solution $w^*$. To have a smooth solution, FPTT learns from the historical information of weight changes by introducing a running mean $\bar{w}_t$ to summarize the historical information of weight evolution: 
\begin{align}
     &w_{t+1} - w_t = \nabla_w(l^{t+1}) - \nabla_w(l^{t}) \\
     \Rightarrow \; &\bar{w}_{t} - w_{t} \sim \nabla_w(l^{t+1}) - \nabla_w(l^{t}) \\
     \Rightarrow \; &\nabla_w(l^{t+1}) - \nabla_w(l^{t}) = \alpha[(\bar{w}_{t} - w_{t}) - (w_{t+1} - w_t)] = \alpha(\bar{w}_{t} - w_{t+1}) \label{eq:w_bar}
\end{align}

From this, the convergence-guaranteed loss function for online update is derived, based on Eq. \eqref{eq:w_bar}. 
\begin{align}
     &\nabla_w(l^{t+1}) - \nabla_w(l^{t}) = \alpha(\bar{w}_{t} - w_{t+1}) \qquad \Leftrightarrow \qquad \nabla_w(l^{t+1}) - \nabla_w(l^{t}) - \alpha(\bar{w}_{t} - w_{t+1}) = 0 \label{eq:l_con}
\end{align}
we define the constraint into the function $f(w_{t+1}) = \nabla_w(l^{t+1}) - \nabla_w(l^{t}) - \alpha(\bar{w}_{t} - w_{t+1})$. We now consider a convex function $F(w)$ which approached its minimum when $f(w_{t+1})=0$; we then have 
\begin{align} 
F (w) & = \int_w f(w) d w \text{ -- searching $w_{t+1}$ over parameter space}\\
& = l^t(w)+\frac{\alpha}{2}\| w - \bar{w}_t - \frac{1}{2\alpha}\nabla_w(l(w_t))\|^2
\end{align}
In this form, Eq~\ref{eq:l_con} is the first order condition for $F(w)$. So, the weight optimization is to minimize the new objective function
\begin{equation}
   w_{t+1} = \underset{w}{\arg\min} \;l^t(w)+\frac{\alpha}{2}\| w - \bar{w}_t - \frac{1}{2\alpha}\nabla_w(l(w_t))\|^2
\end{equation}

\end{document}